\documentclass{isprs}
\usepackage{subfigure}
\usepackage{setspace}
\usepackage{geometry} 
\usepackage{epstopdf}
\usepackage[labelsep=period]{caption}  
\usepackage{float}
\usepackage{microtype} 
\usepackage{icomma}
\usepackage[range-phrase={\:to\:}]{siunitx}
\usepackage{xspace}
\usepackage{amssymb}
\usepackage{amsmath}
\usepackage{url}
\usepackage{natbib}
\usepackage{enumitem}
\usepackage{pgfplots}
\usepackage{etoolbox}
\usepackage{hyperref}
\usepackage{textcomp}
\usepackage{hhline}
\usepackage{ucs}
\usepackage[utf8]{inputenc}
\usepackage[T1]{fontenc}

\usepackage{multirow}
\usepackage[normalem]{ulem}
\useunder{\uline}{\ul}{}

\usepackage{float}
\usepackage{graphicx}
\usepackage{color}
\DeclareGraphicsExtensions{.pdf,.eps,.png,.jpg,.mps}
\graphicspath{{figures/}}

\usepackage{tikz}
\usetikzlibrary{arrows,shapes.geometric,positioning,calc}
\definecolor{KITMint}{RGB}{0,150,130}
\definecolor{KITRed}{RGB}{162,34,35}
\definecolor{KITBlue}{RGB}{70,100,170}
\definecolor{KITOrange}{RGB}{223,155,27}
\definecolor{KITGelb}{RGB}{252,229,0}
\definecolor{KITGruen}{RGB}{140,180,60}

\definecolor{VoxelCeiling}{RGB}{255,0,0}
\definecolor{VoxelFloor}{RGB}{0,255,0}
\definecolor{VoxelWall}{RGB}{200,200,200}
\definecolor{VoxelWallOpening}{RGB}{100,149,237}
\definecolor{VoxelInteriorObject}{RGB}{105,105,105}
\definecolor{Yellow}{RGB}{255,255,0}

\usepackage{fontawesome}
\usepackage{booktabs}

\begin{document}

\title{Voxel-based Indoor Reconstruction From HoloLens Triangle Meshes}

\author{
 P. Hübner, M. Weinmann, S. Wursthorn
}

\address
{
 Institute of Photogrammetry and Remote Sensing, Karlsruhe Institute of Technology (KIT), Karlsruhe, Germany\\ (patrick.huebner, martin.weinmann, sven.wursthorn)@kit.edu\\
}


\abstract{
    Current mobile augmented reality devices are often equipped with range sensors.
The Microsoft HoloLens for instance is equipped with a Time-Of-Flight (ToF) range camera providing coarse triangle meshes that can be used in custom applications.
We suggest to use the triangle meshes for the automatic generation of indoor models that can serve as basis for augmenting their physical counterpart with location-dependent information.
In this paper, we present a novel voxel-based approach for automated indoor reconstruction from unstructured three-dimensional geometries like triangle meshes.
After an initial voxelization of the input data, rooms are detected in the resulting voxel grid by segmenting connected voxel components of ceiling candidates and extruding them downwards to find floor candidates.
Semantic class labels like 'Wall', 'Wall Opening', 'Interior Object' and 'Empty Interior' are then assigned to the room voxels in-between ceiling and floor by a rule-based voxel sweep algorithm.
Finally, the geometry of the detected walls and their openings is refined in voxel representation.
The proposed approach is not restricted to Manhattan World scenarios and does not rely on room surfaces being planar.
}


\maketitle

\section{INTRODUCTION}
\label{sec:introduction}

Current head-worn Augmented Reality (AR) devices like the Microsoft HoloLens\footnote{\url{https://www.microsoft.com/en-us/hololens}} hold great potential for enriching indoor environments with the in-situ visualisation of location-dependent information, e.g. from digital building information models \citep{huebner_et_al_2018b}.
While suchlike building models are currently gaining in prevalence with the increasing use of Building Information Modeling (BIM) techniques in the planning and construction stages of building projects \citep{jung_lee_2015}, already existing, older buildings frequently lack such a digital representation that could be used in such indoor AR scenarios \citep{lu_lee_2017}. 

Mobile Indoor AR devices are however often equipped with range sensors to facilitate 1) tracking and relocalisation within indoor environments and 2) convincing placement and interaction of virtual content with the physical surrounding.
The HoloLens for instance is equipped with a Time-of-Flight (ToF) range camera providing range images and preprocessed triangle meshes that can be used in custom applications \citep{huebner_et_al_2019, khoshelham_et_al_2019}.
We suggest to use these data for the automatic generation of digital models of indoor environments that can serve as basis for augmenting their physical counterpart with location-dependent informative content in an indoor AR setting.

While the automated reconstruction of indoor building models from unstructured three-dimensional geometries is an active field of research \citep{ma_liu_2018}, prevalent approaches use dense point clouds acquired by LiDAR sensors or range cameras.
To the best of our knowledge, there is currently no attempt on indoor reconstruction using sparse triangle meshes as provided by the HoloLens system.

In this paper, we present a novel indoor reconstruction approach for automatically deriving indoor building models as voxel representation from sparse triangle meshes.
The proposed approach is not restricted to Manhattan World scenarios and does not rely on room surfaces being planar.
It can furthermore be easily transferred to using input data in the form of dense point clouds in the scope of future work. 

After briefly summarising related work in Section 2
, we elaborately describe our voxel-based method for the reconstruction of indoor environments from sparse triangle meshes in Section 3
and the applied evaluation procedure.
After presenting evaluation results in Section 4
and discussing them in Section 5
, we provide concluding remarks and suggestions for further research in Section~6.
\section{RELATED WORK}
\label{sec:related_work}

Recently, a range of notable work on indoor reconstruction was published which is briefly summarised in this section.
\citet{tran_khoshelham_2019} for instance proposed to detect rooms in LiDAR point clouds via a space subdivision approach by RANSAC planes, first horizontally for separating storeys and then vertically per storey.
In each storey, the resulting cell complex generated by the intersecting planes is arranged to the most probable composition of rooms by a stochastic approach, where the likelihood of a cell being part of a room is determined by the amount of points constituting its surface. Points obstructing room interior space are penalised. 

\citet{yang_et_al_2019} proposed to detect rooms in a cell complex of intersecting lines by an energy minimisation approach.
Wall surfaces constituting the cell complex are derived in this case by detecting and refining closed contours in 2D sections shortly below ceiling height, where the amount of furniture can be expected to be minimal.
This approach is capable of reconstructing rooms with horizontally curved walls.

\citet{ochmann_et_al_2019} focused on indoor reconstruction relying on a RANSAC-based for plane detection. 
Rooms are segmented from the detected planes by Markov clustering based on the mutual visibility of plane patches. 
The detected room surface planes are subsequently intersected in 3D space to reconstruct volumetric wall and slab objects via an integer linear programming approach.

\citet{nikoohemat_et_al_2018b} presented an approach to reconstruct indoor models from point clouds captured by mobile LiDAR-based indoor mapping systems while also incorporating trajectory information e.g. for separating storeys and stairwells.
Planar surface patches are detected and successively merged depending on coplanarity and distance.
The resulting plane patches are arranged in an adjacency graph and subjected to a rule-based process to reconstruct and refine room surface geometry.
Wall openings are distinguished from occlusion-caused absence of points by a voxel-based ray-tracing approach emenating from the trajectory and by checking if the trajectory itself passes through an opening.
Rooms are discerned by a voxel-based space partitioning, again incorporating trajectory information.

\citet{xie_et_al_2019} also focused on detecting planar patches and subsequently refining them by merging.
Then vertical patches are selected based on their normal direction as wall surfaces and contour lines are extracted in 2D sections in a similar way as proposed by \citet{yang_et_al_2019}.
Here, however, multiple contours are extracted over the whole height range of the room to account for vertical changes in wall geometry like protrusions or recesses.
The refined contours are then assembled to 3D room surfaces.

\citet{diaz-vilarino_et_al_2019} presented an indoor reconstruction approach, that also relies on region growing for extracting planar surfaces.
The planes are subsequently classified in obstacles and room surfaces, with the latter being further refined by intersecting the individual planes with each other.
For each wall surface, rectangular openings are detected by applying the Generalized Hough Transformation on 2D raster representations of the walls.
The reconstructed indoor models are further used in an indoor pathfinding framework.

Like \citet{nikoohemat_et_al_2018b}, \citet{flikweert_et_al_2019} also use mobile LiDAR point clouds while also taking the trajectory into account as additional source of information.
Wall surfaces are detected in the point cloud by applying the Medial Axis Transform.
Doors are then detected in voxelized walls with regard to the voxelized trajectory.
Rooms are finally segmented by region growing among floor voxels while stopping on floor voxels within the detected doors.

\citet{gorte_et_al_2019} also use a 3D voxel representation of the input data for floor detection.
They however do not reconstruct complete indoor models but focus on detecting navigable floor space for indoor path-finding to building exits.
Floor voxels are detected as non-empty voxels with a certain amount of vertical and horizontal free space above them. 
Region growing with a threshold on the maximum height difference between neighbouring floor voxels generates connected components of navigable floor voxels extending over stairs.
Here, indoor voxel models prove to be a suitable representation for distance calculation for navigation tasks.
\section{METHODOLOGY}
\label{sec:method}


    In this section, we present our novel method for reconstructing voxel models of indoor environments from unstructured 3D data with normal directions.
    After stating necessary data preparation steps, we provide an in-depth explanation of our proposed reconstruction algorithm.
    Subsequently, we elaborate on our evaluation procedure.
    
    The proposed approach is summarised in \autoref{fig:workflow} with Table \autoref{tab:voxelclasses} detailing the respective voxel colours. 
    3D data of indoor environments are transforms to a voxel representation and subsequently a model of the respective indoor environment is reconstructed in voxel space.
    To this aim, voxels are assigned to rooms and semantic classes.
    While the method is applicable for unstructured 3D data with absolute normals and thus could also be applied to point clouds provided they have normals, we focus in this study on triangle meshes captured with the Microsoft HoloLens.
    
        \subsection{Data Preparation}
        \label{sec:method_dataPreparation}
        
            While our proposed method is not restricted to Manhattan World scenarios, where all room surfaces have to be aligned with the coordinate axes, the results are cleaner and visually more agreeable when planar wall surfaces are aligned with the coordinate axes.
            Thus, it can make sense to align the model along the coordinate axes as a preprocessing step.
            This is, however, optional and not necessary for the reconstruction method presented here to work.
            Furthermore, our method does not rely on wall surfaces to be planar but can also deal with curved wall surfaces.
            
            For now, however, we presuppose, that walls are vertical.
            Vertically slanted walls with an inclination of more than \SI{45}{\degree} are valid though.
            These are reconstructed as ceiling surfaces.
            We furthermore assume that the upward direction is known and corresponds to one of the coordinate axes.
            
            As a preliminary processing step, the input dataset is converted to a voxel representation.
            To this aim, a certain range of space encompassing the input data is subdivided by a three-dimensional grid of cubical, non-overlapping cells of uniform size, i.e. voxels.
            Each voxel intersecting a geometric primitive of the input data (i.e. a point of a point cloud or a triangle of a triangle mesh as is the case here) is marked as non-empty.
            These non-empty voxels are classified according to the normal directions of the intersecting geometric primitives.
            If the majority of the normal vectors of intersecting primitives points upwards or downwards within a range of $\pm$\SI{45}{\degree}, the voxel is classified as 'Normal Up' or 'Normal Down', respectively.
            Otherwise, non-empty voxels are classified as 'Normal Horizontal'.
            
            Voxels can thus be characterised as having one of four possible states: 'Empty', 'Normal Up', 'Normal Down' or 'Normal Horizontal'.
            Hence, the voxel representation can be stored memory-efficiently as a three-dimensional array of byte values.
            The voxel size as parameter of the voxelization process is set to \SI{5}{cm} in the scope of this paper, a reasonable value in consideration of the typical dimensions of indoor environments and the resolution of the HoloLens triangle meshes.
            The same voxel size is used e.g. by \citet{gorte_et_al_2019} for the automated extraction of navigable space in indoor environments.
            
            The resulting voxel grid serves as input for the reconstruction algorithm presented in the following sections.
            The aim of the proposed algorithm is to segment this voxel grid in rooms and to classify voxels belonging to a room as 'Ceiling', 'Floor', 'Wall', 'Wall Opening', 'Empty Interior' or 'Interior Object'.
            
        \subsection{Room Detection}
        \label{sec:method_roomDetection}
        
            In a first processing step, rooms are detected in the voxel model by segmenting ceiling candidates.
            In subsequent refinement steps, ceilings and floors are reconstructed as voxel representation for each room.
        
            \subsubsection{Ceiling Detection}
            \label{sec:method_roomDetection_ceilingDetection}
        
                Initially, ceiling segments are detected as seeds for room candidates.
                This is achieved by segmenting 'Normal Down' voxels to connected components based on a 3D-26-neighbourhood via 3D region growing.
                Beforehand, a rule-based voxel sweep is performed by iterating section-wise from bottom to top through the voxel grid along the upward direction and switching 'Normal Down' voxels that have a 'Normal Horizontal' voxel directly below them to 'Normal Horizontal' themselves.
                This ensures that the ceiling segments resulting from the region growing algorithm represent distinct room candidates divided by walls.
                
                Ceiling segments are discarded as room candidates if they have a horizontal coverage of less than \SI{0.5}{\metre\squared}.
                In our algorithm, parameters like this one are generally set as metric values in order to be independent of the applied resolution of the voxel grid.
                The respective values are chosen in consideration of typical dimensions of indoor environments.
                
            \subsubsection{Ceiling Refinement}
            \label{sec:method_roomDetection_ceilingRefinement}
            
                \begin{table}
    \caption{Color schemes for voxel classes.}
    \begin{subtable}{}
        \label{tab:voxelclasses_normal}
        \begin{tabular}{cl}
            \multicolumn{2}{c}{a) Normal Direction} \\
            \toprule
    		Color & Class Label \\ \midrule
    		\textcolor{VoxelCeiling}{\faCube} & Normal Down \\
    		\textcolor{VoxelFloor}{\faCube} & Normal Up \\
    		\textcolor{VoxelWall}{\faCube} & Normal Horizontal \\
    		\bottomrule
    		\multicolumn{2}{c}{} \\
            \multicolumn{2}{c}{} \\
    	\end{tabular}
    \end{subtable}
    \begin{subtable}{}
        \label{tab:voxelclasses_reconstruction}
        \begin{tabular}{cl}
    		\multicolumn{2}{c}{b) Indoor Reconstruction} \\
            \toprule
    		Color & Class Label \\ \midrule
    		\textcolor{VoxelCeiling}{\faCube} & Ceiling \\
    		\textcolor{VoxelFloor}{\faCube} & Floor \\
    		\textcolor{VoxelWall}{\faCube} & Wall \\
    		\textcolor{VoxelWallOpening}{\faCube} & Wall Opening \\
    		\textcolor{VoxelInteriorObject}{\faCube} & Interior Object \\ \bottomrule
    	\end{tabular}
    \end{subtable}
	\label{tab:voxelclasses}
\end{table}
                
                The remaining ceiling candidates are subsequently refined.
                In this refinement process, horizontal holes in the voxel segments representing ceilings are detected.
                These holes are eventually closed later on if they satisfy certain criteria trying to ensure that only holes are being closed, that are caused by incomplete acquisition of the geometric primitives the voxel representation is derived from, e.g. caused by occlusion.
                Actual holes in ceiling surfaces caused e.g. by columns or corners pointing convexly inside the room, on the other hand, should not be closed.
                
                For the detection of holes in ceiling segments, the ceiling segments are transformed to a horizontal two-dimensional pixel grid covering the whole horizontal extension of the respective ceiling segment. 
                Pixels are marked as non-empty if a voxel of the ceiling segment occupies its position.
                The ceiling pixel is assigned the height in voxel grid integer coordinates of the lowest ceiling segment voxel occupying its position.
                
                Hole pixels are then detected within this two-dimensional grid by searching for empty pixels that are in-between two non-empty pixels along the four main directions resulting from a 2D-8-neighbourhood.
                Hole pixels need not be directly neighbouring its enclosing non-empty pixels, i.e. holes can have an extent of multiple pixels.
                The detected hole pixels are subsequently segmented to hole segments based on a 2D-4-neighbourhood.
                
                Furthermore, each hole pixel is assigned a height value in integer voxel grid coordinates.
                This height value is determined by 2D ray tracing from ceiling pixels that are positioned in 2D-4-neighbourhood of hole pixels across the hole in the four main directions resulting from a 2D-8-neighbourhood.
                If a 2D ray ends in another ceiling pixel, the height of hole pixels along the ray is interpolated linearly between the height values of the two ceiling pixels.
                Otherwise, the height of hole pixels is set directly from the ceiling pixel the ray started from.
                The height value of each hole pixel is kept as a moving average over all rays and subsequently rounded to an integer value.
                Finally, a smoothing of the height values is applied across all hole pixels belonging to a common hole segment.
                

\pgfdeclarelayer{bg}
\pgfsetlayers{bg,main}

\tikzstyle{preparation} = [rectangle, rounded corners, minimum width=5.2cm,minimum height=0.7cm, text centered, draw=red, fill=red!20]
\tikzstyle{roomdetection} = [rectangle, rounded corners, minimum width=5.2cm,minimum height=0.7cm, text centered, draw=blue, fill=blue!20]
\tikzstyle{voxelclassification} = [rectangle, rounded corners, minimum width=5.2cm,minimum height=0.7cm, text centered, draw=orange, fill=orange!20]
\tikzstyle{refinement} = [rectangle, rounded corners, minimum width=5.2cm,minimum height=0.7cm, text centered, draw=green, fill=green!20]

\tikzstyle{startstop} = [rectangle,inner sep=0pt,text centered,minimum width=5.2cm,minimum height=0.7cm]

\tikzstyle{arraow} = [thick,->,>=stealth]
	
\begin{figure}[H]
\centering
	\begin{tikzpicture}[%
		node distance = 0.3cm,
		every node/.style={font=\sffamily},
		align=center,
		]
		
		\node[startstop] (mesh) {Input mesh\\\includegraphics[width=0.6\columnwidth]{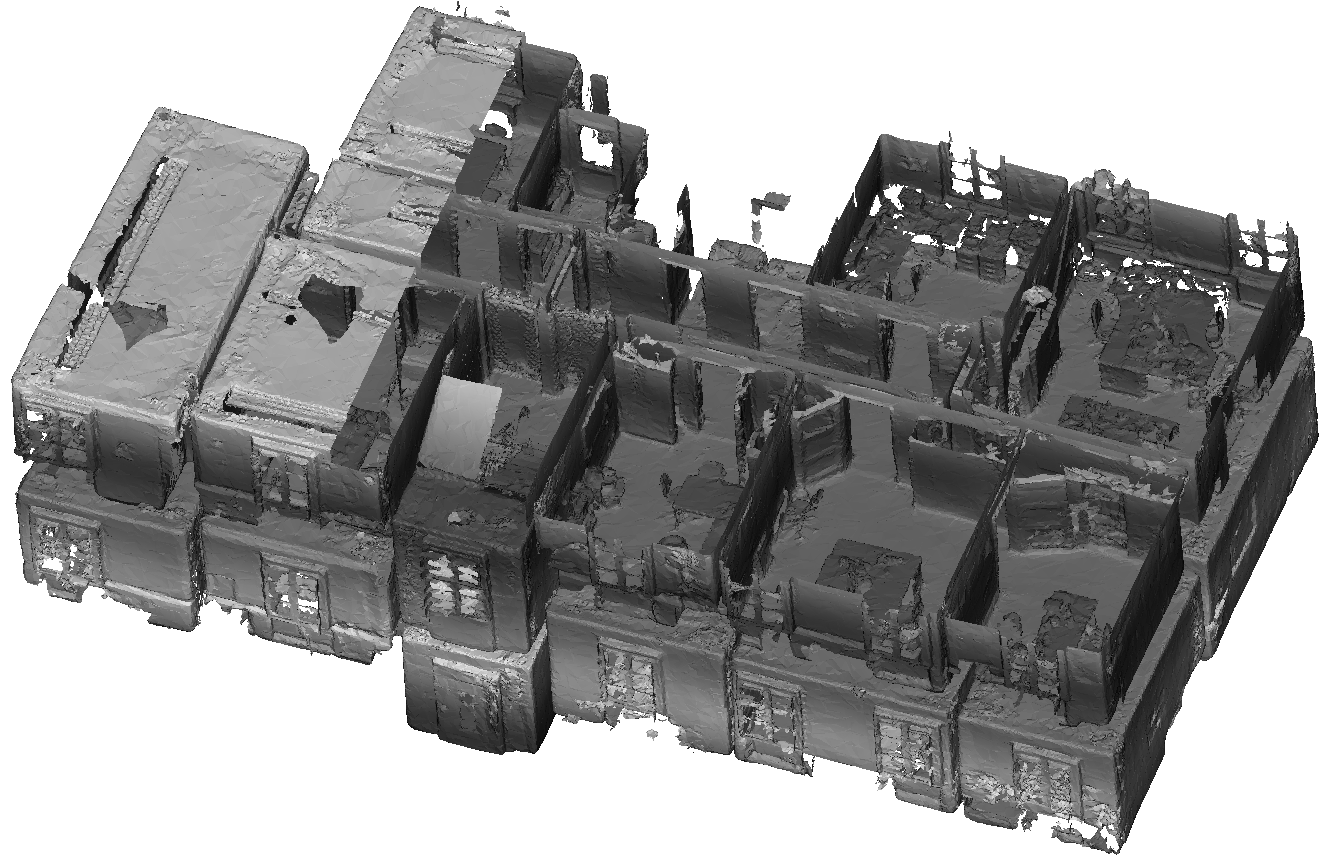}};
		
		\node[preparation,below=of mesh] (voxelization) {Voxelization \& Normal Classification\\\includegraphics[width=0.4\columnwidth]{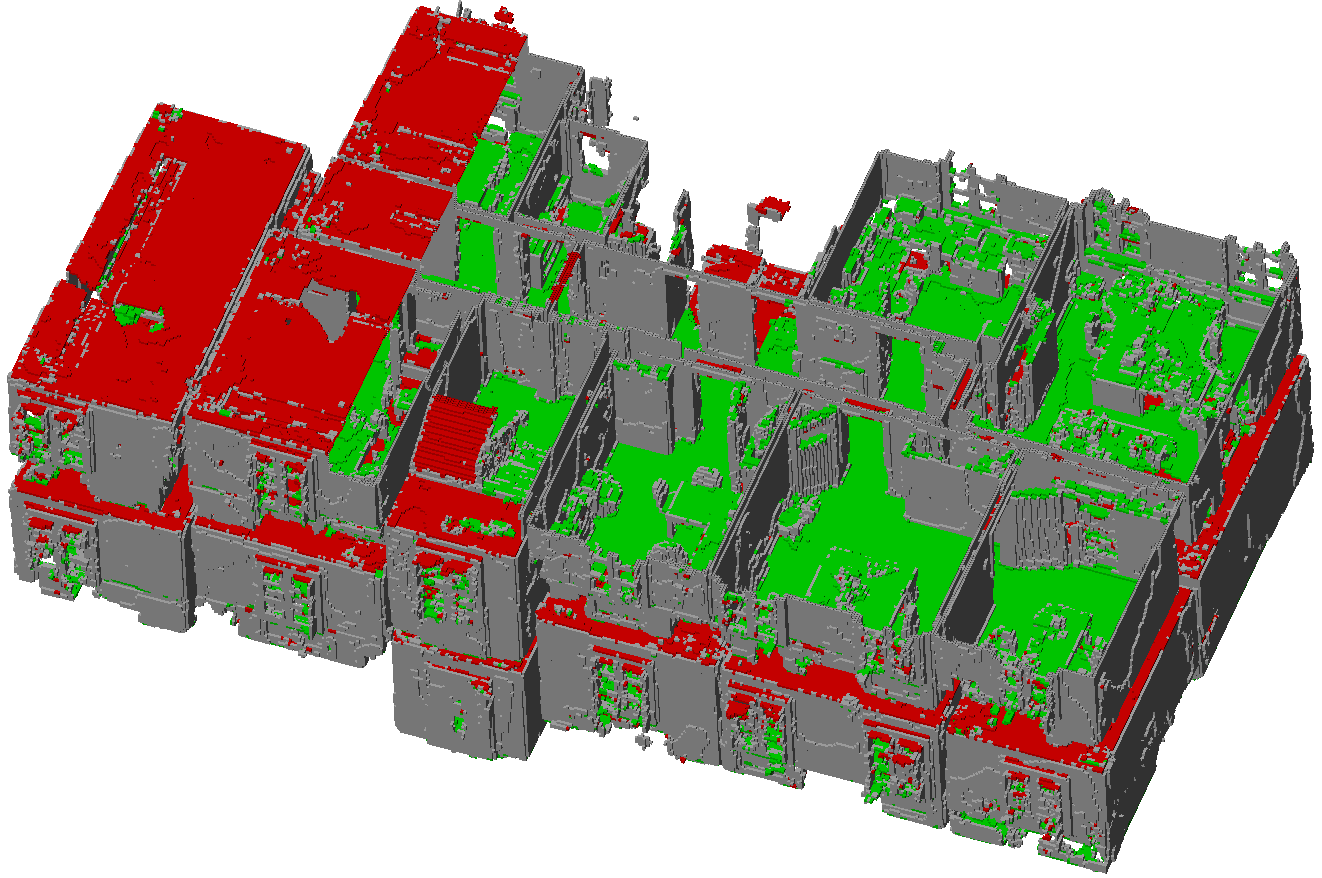}};
				
		\node[roomdetection,below=of voxelization] (ceilingdetection) {Ceiling Detection\\\includegraphics[width=0.4\columnwidth]{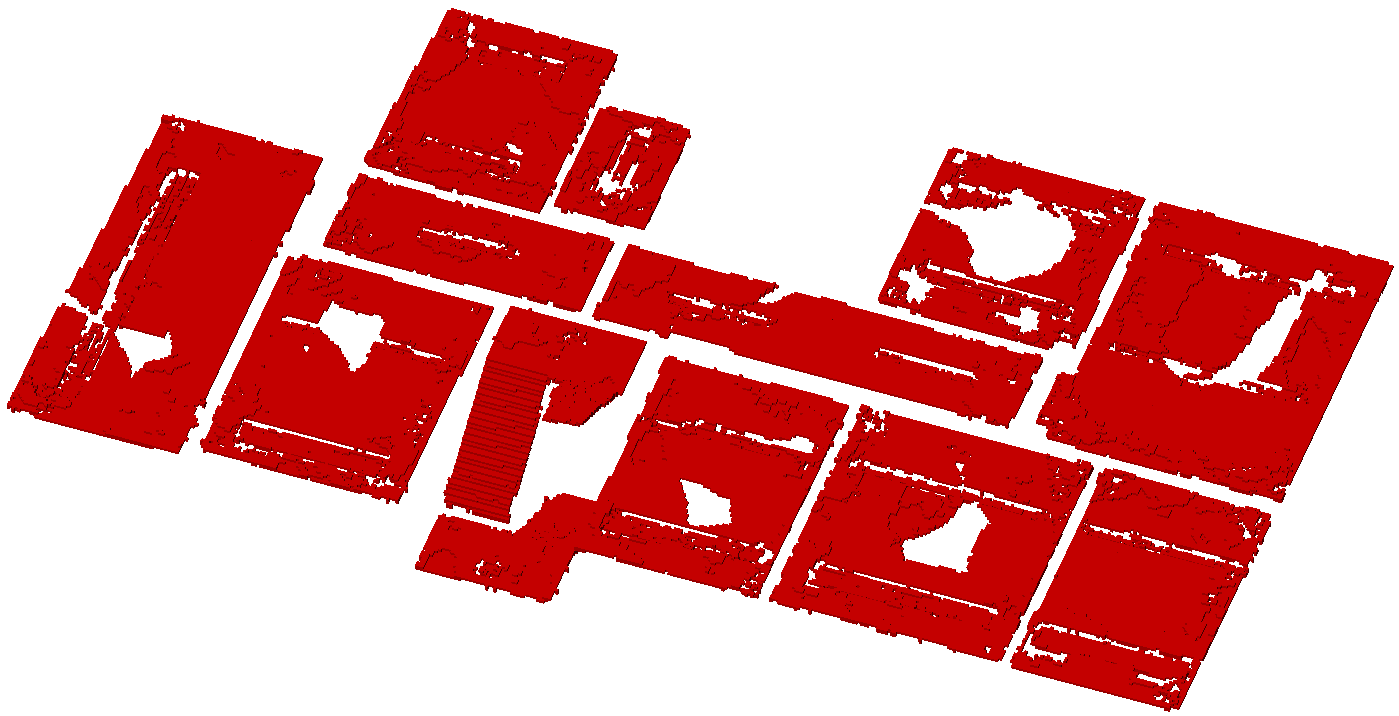}};
		\node[roomdetection,below=of ceilingdetection] (ceilingrefinement) {Ceiling Refinement};
		\node[roomdetection,below=of ceilingrefinement] (floordetection) {Floor Detection};
		\node[roomdetection, below=of floordetection] (ceilingfloorfinalization) 
		{Ceiling \& Floor Finalization\\\includegraphics[width=0.4\columnwidth]{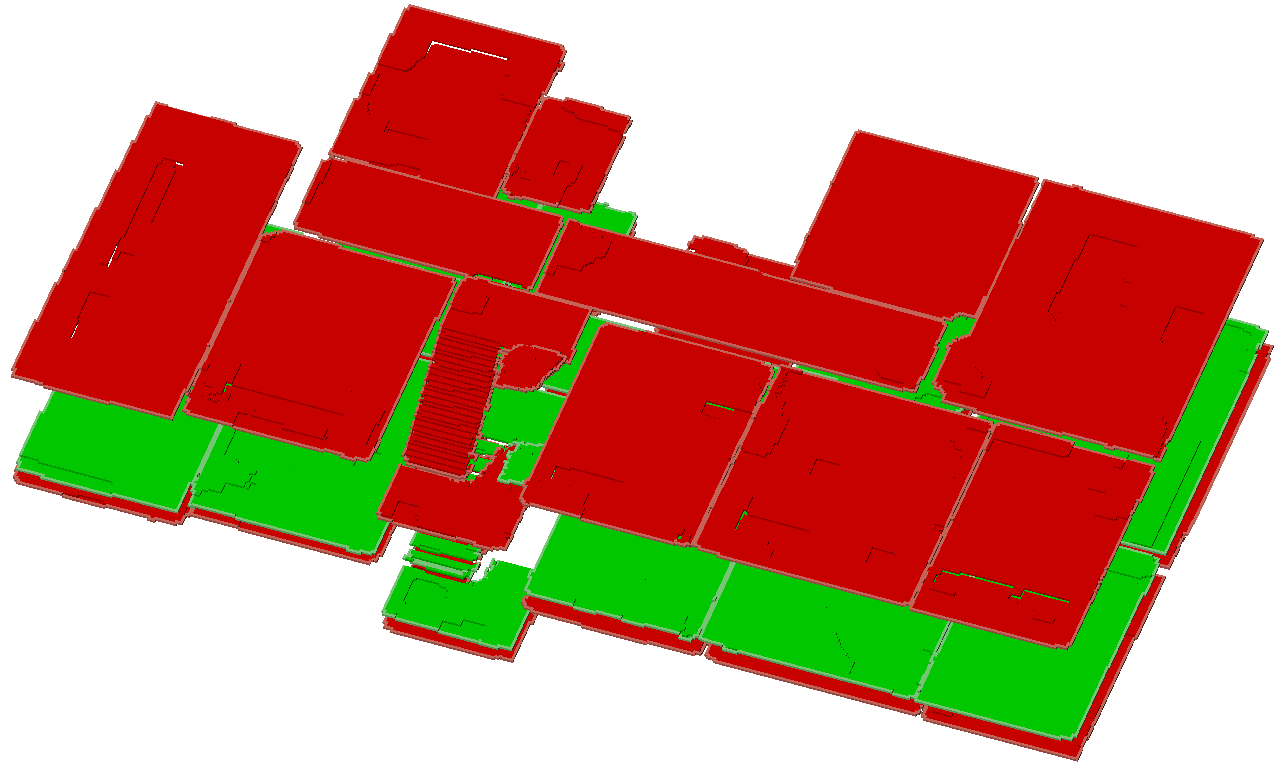}};
		
		\node[voxelclassification, below=of ceilingfloorfinalization] (voxelclassification) 
		{Voxel Classification Sweep\\\includegraphics[width=0.4\columnwidth]{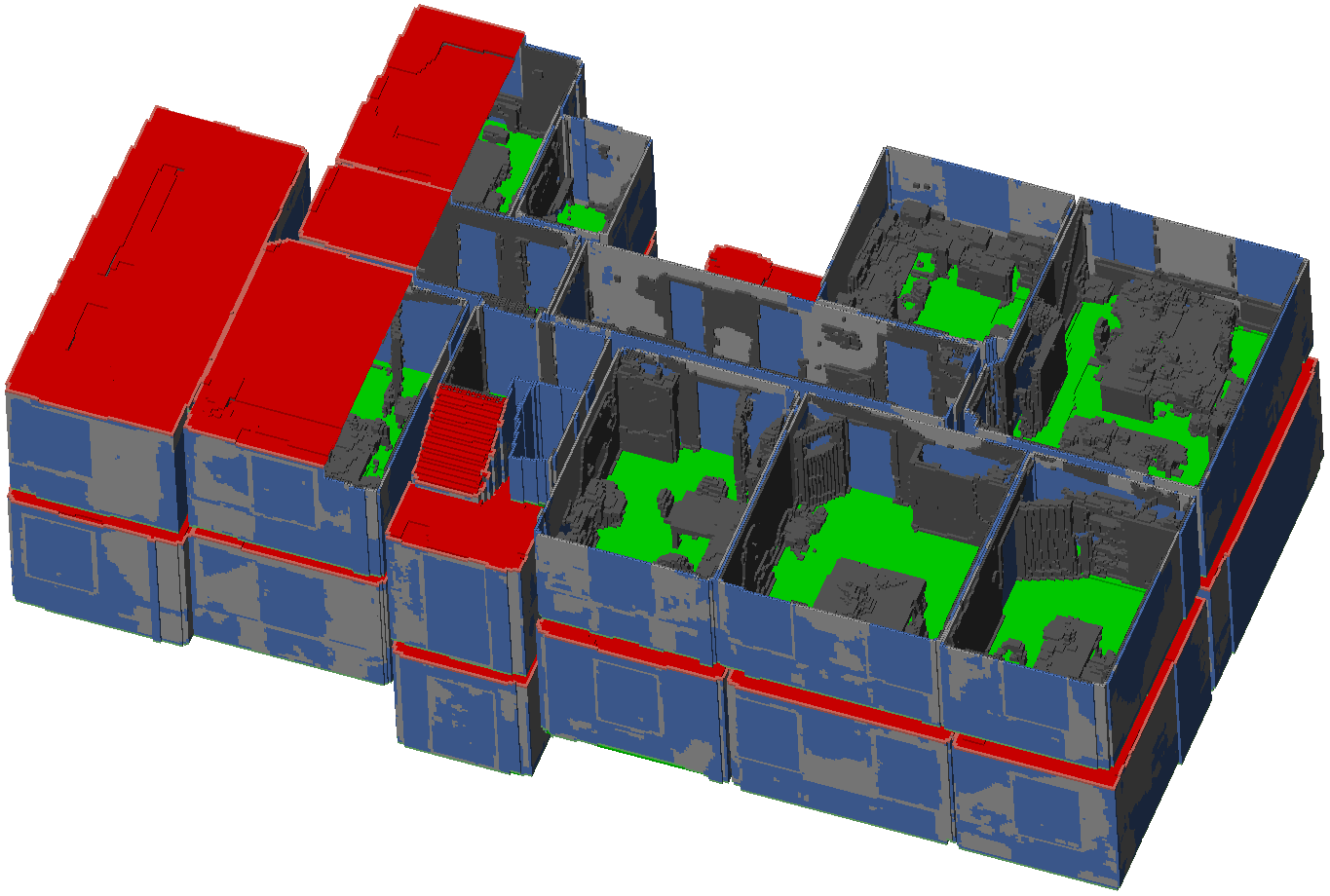}};
		
		\node[refinement,below=of voxelclassification] (wallgeomertyrefinement) {Wall Geometry Refinement};
		\node[refinement,below=of wallgeomertyrefinement] (wallopeningrefinement) {Wall Opening Refinement};
		
		\node[startstop, below=of wallopeningrefinement] (result) {\includegraphics[width=0.6\columnwidth]{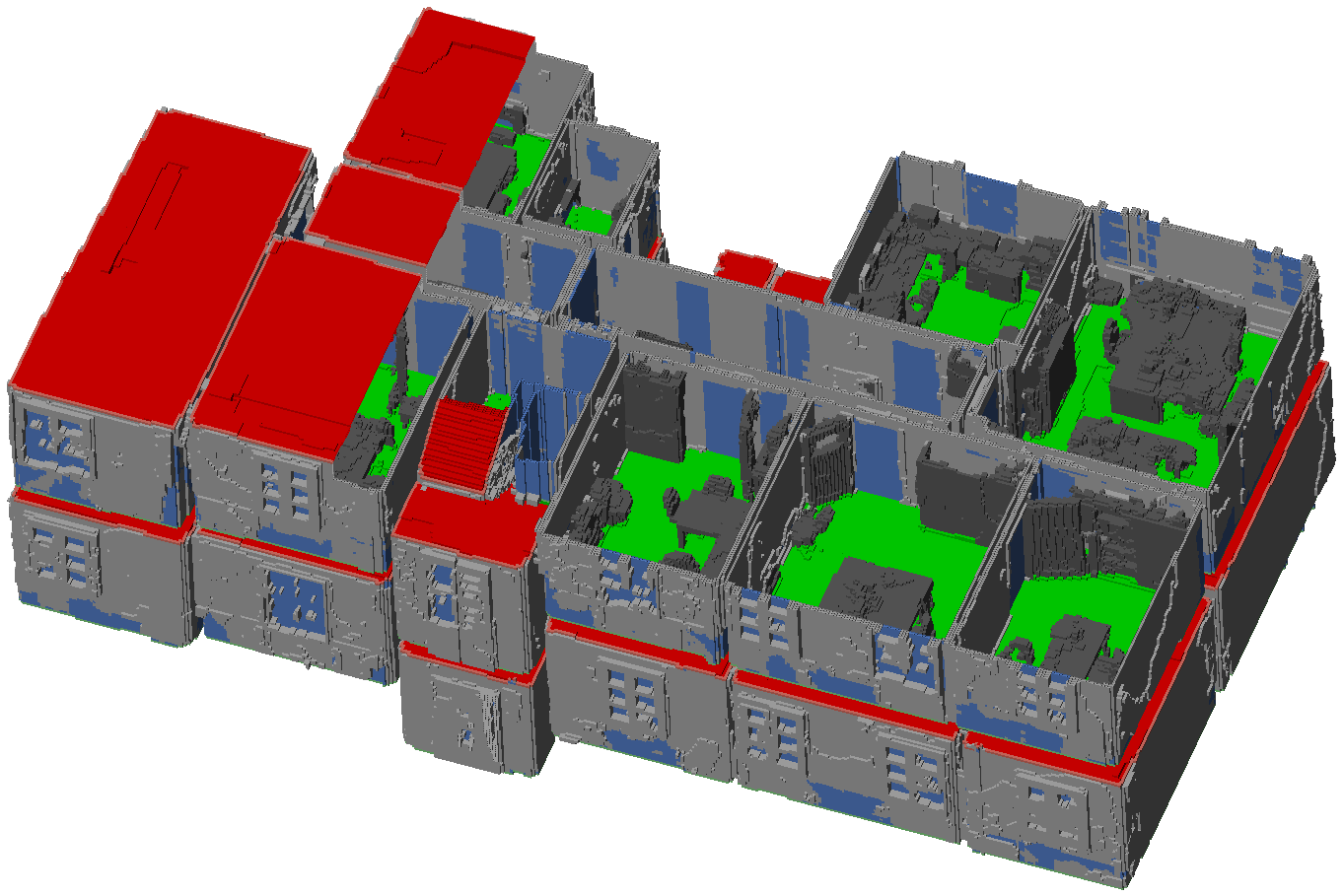}\\Result: reconstructed voxel model};
		
		\draw[dashed,draw=lightgray,thick] let \p1=(voxelization),\p2=(ceilingdetection),\p3=(ceilingdetection.north west),\p4=(ceilingdetection.north east) in
			(\x3-1cm,-4.9) -- (\x4+1cm,-4.9);
		\draw[dashed,draw=lightgray,thick] let \p1=(ceilingfloorfinalization),\p2=(voxelclassification),\p3=(voxelclassification.north west),\p4=(voxelclassification.north east) in
			(\x3-1cm,-12.3) -- (\x4+1cm,-12.3);
		\draw[dashed,draw=lightgray,thick] let \p1=(voxelclassification),\p2=(wallgeomertyrefinement),\p3=(wallgeomertyrefinement.north west),\p4=(wallgeomertyrefinement.north east) in
			(\x3-1cm,-15.35) -- (\x4+1cm,-15.35);
		
		\begin{pgfonlayer}{bg} 
			\draw[thick,->] (0,0) -- (result.north);
		\end{pgfonlayer}
		
		\node[rotate=90,scale=1.2,text=red] at (-3.2,-3.3) {Data Preparation};
		\node[rotate=90,scale=1.2,text=blue] at (-3.2,-8.5) {Room Detection};
		\node[rotate=90,scale=1.2,text=orange] at (-3.2,-13.85) {Voxel\\Classification};
		\node[rotate=90,scale=1.2,text=green!50!black] at (-3.2,-16.52) {Voxel Model\\Refinement};
			
	\end{tikzpicture}
	\caption{
	    Processing workflow from input mesh to final reconstructed voxel model. The colour scheme of the normal classification is described in Table \ref{tab:voxelclasses_normal}. The other depictions are coloured according to Table \ref{tab:voxelclasses_reconstruction}. A part of the ceiling is removed for visibility.
    }
	\label{fig:workflow}
\end{figure}
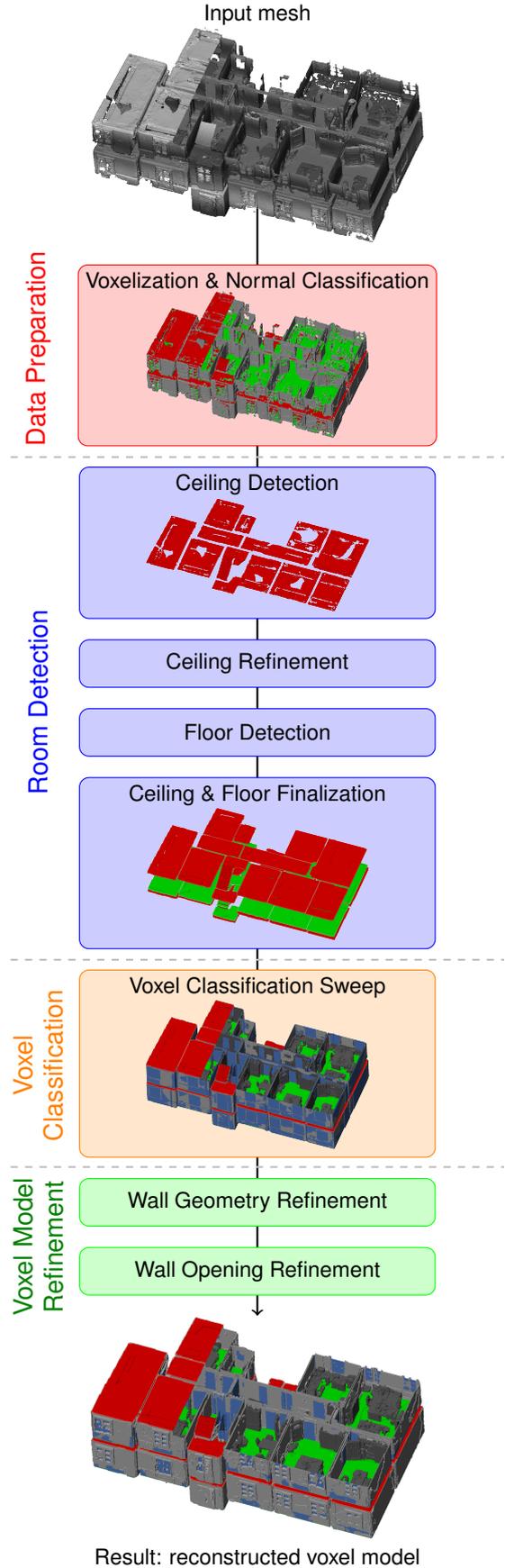
                
            \subsubsection{Floor Detection}
            \label{sec:method_roomDetection_floorDetection}
            
                Once all hole pixels have been assigned a height value, it is decided per hole segment if the hole can be closed, i.e. if the voxels corresponding to hole pixels and their respective height values should be included as ceiling voxels in the respective ceiling segment.
                To make this decision, knowledge about the supposed vertical extent of the room under the respective ceiling voxel is necessary.
                Hence, for each voxel correponding to a ceiling or hole pixel and its respective height, a corresponding floor voxel candidate needs to be found.
                
                To this aim, another two-dimensional pixel grid of the same extent as the respective ceiling pixel grid is created for each ceiling voxel segment.
                The voxel grid is traced through downwards from each ceiling or hole voxel position.
                If a 'Normal Up' voxel is encountered before reaching the bottom of the voxel grid or a 'Normal Down' voxel belonging to another ceiling segment, the respective floor pixel grid position is marked as floor candidate and assigned the height value of the 'Normal Up' voxel.
                
                These floor candidate pixels are subsequently segmented in floor candidate pixel segments based on a 2.5D-8-neighbourhood per 2D floor grid.
                The 2.5D-8-neighbourhood is realised by 2D region growing with a threshold of \SI{18}{cm} (a common height of stair steps) as maximally allowed height difference between neighbouring floor candidate pixels.
                
                Floor candidate segments with a horizontal extent of less then \SI{0.5}{\metre\squared} are discarded, unless there are no floor candidate segments with a horizontal extent larger than this in the respective floor pixel grid.
                From the remaining floor candidate segments, the segment containing the lowest height is selected as final floor for the respective room.
                All pixels belonging to other segments are set to 'Empty'.
                
                The selected value of \SI{18}{cm} as maximum height difference on floors proved to result in floor segments that can extend over stairs and ramps while mostly avoiding to spread over to surfaces on top of furniture like tables or chairs.
                
                For each floor pixel grid, all empty pixels corresponding to a ceiling pixel or hole pixel in the respective corresponding ceiling pixel grid need to have a height value assigned.
                Their height is determined via the same procedure as the height of the hole pixels in the ceiling pixel grids as detailed beforehand.
                
            \subsubsection{Ceiling And Floor Finalisation}
            \label{sec:method_roomDetection_ceilingFloorFinalisation}
                
                On the basis of the floor reconstruction, the decision which ceiling holes are to be closed can now be made by analysing the voxels below hole voxels in the height range restricted by the height values in the respective floor grid pixels.
                If a horizontal hole extent overlaps with another already reconstructed room within this height range, it cannot be closed.
                If this is not the case and at least \SI{75}{\percent} of the horizontal hole extent is occupied by a non-empty voxel in the input voxel grid within the height range between ceiling and floor, the hole is to be closed.
                Otherwise, the hole can potentially still be closed, unless the vertical extent of the border between hole and room is filled to at least \SI{75}{\percent} in the input voxel grid, indicating the presence of a wall.
                
                At this state, ceilings and floors of all detected rooms are determined as connected voxel components with each ceiling voxel having a corresponding floor voxel somewhere below it.
                Subsequently, a smoothing process is applied to the height of ceilings and floors per room.
                Furthermore, wall voxels are initialized along the border of ceilings as a closed contour of voxels neighbouring the ceiling segment on the outside.
            
        \subsection{Voxel Classification}
        \label{sec:method_classification}
            
                During the complete reconstruction process detailed here, the state of the voxel grid is stored in a three-dimensional array of integer arrays.
                Under certain circumstances, our algorithm allows for voxels to belong to two different classes at once for the same room as well as to belong to multiple rooms at once with different classes.
                For instance, a voxel can be classified as 'Ceiling' as well as as 'Wall' at once for the same room to represent the fact, that the edge between the actual ceiling and wall surfaces runs through this voxel.
                This is the case for the wall voxels initialized along the contours of ceilings.
                Similarly, voxels can belong simultaneously to the classes 'Floor' and 'Wall'.
                
                Furthermore, voxels can belong to more than one room.
                This is the case, if the surfaces of neighbouring rooms are covered by the same voxel.
                The frequency of this occurrence depends on the voxel size as well as on the width of walls and horizontal slabs of the buildings represented by the data.
                In those cases where a voxel belongs to more than one room, its respective classes can only occur in certain specific combinations.
                While a voxel can e.g. represent the walls of two neighbouring rooms or ceiling for one room and floor for another directly above it, it for example cannot represent floor for more than one room at once.
                Also voxels belonging to the room interior (classes 'Empty Interior' and 'Interior Objects') can only belong to one room, as rooms can only share voxels along their borders.
                
                For classifying voxels besides the already reconstructed floors and ceilings, the voxel grid is traversed section-wise from top to bottom, while voxels are assigned to rooms and classes depending on the state of the voxel above it.
                In doing so, all voxels below a ceiling voxel get assigned to the same room the ceiling voxel belongs to, until a floor voxel of the respective room is encountered.
                When the ceiling voxel does not also have the class label 'Wall' for the same room, the voxels between it and the floor will be assigned to the classes 'Empty Interior' or 'Interior Object', depending on the respective voxel being empty or not in the input data.
                Voxels below ceiling voxels that are simultaneously wall voxels of the same room, on the other hand, are classified as 'Wall' or 'Wall Opening' based on the same criterion.
                When walls finally hit their corresponding floor voxel, that voxel gets assigned the class label 'Wall' additionally to it being classified as 'Floor'.
            
        \subsection{Voxel Model Refinement}
        \label{sec:method_refinement}
        
            After applying the voxel sweep algorithm described previously, voxels have assigned class labels and room affiliations. 
            The resulting indoor voxel model is further improved in subsequent refinement sweeps.
            
            Preliminarily, horizontal two-dimensional normal directions in integer voxel grid coordinates are determined pointing towards the room interior for all wall voxels (class 'Wall' or 'Wall Opening') based on a 2D-4-neighbourhood.
            As is the case with class labels, these normal directions are managed on a per-room basis.
            Thus, a voxel can have two sets of normal directions if it is a wall voxel neighbouring two rooms.
        
            \subsubsection{Wall Geometry Refinement}
            \label{sec:method_refinement_wallGeometry}
                
                In a first refinement pass, missing wall sections are detected and completed by searching for voxels belonging to the room interior (classes 'Empty Interior' and 'Interior Objects') that are horizontally neighbouring voxels that do not belong to the same room based on a 2D-4-neighbourhood.
                These neighbouring voxels not belonging to the same room can either belong to other rooms or no room at all.
                
                Such situations can occur on discontinuities in height of horizontally neighbouring ceiling or floor voxels.
                This can occur e.g. on vertical windows in slanted ceilings or on the floor where large discontinuities in height are possible despite of the threshold on height difference of neighbouring voxels in the region growing of floor segments. 
                This can happen, when parts of the floor on different height levels are connected by stairs or ramps (e.g. in the case of staircases or platforms like stages).
                In these cases, interior voxels are converted to wall voxels (class 'Wall' or 'Wall Opening' depending on the voxel being empty in the input data) or to ceiling or respectively floor depending on the height of the discontinuity being larger or smaller than the aforementioned threshold of \SI{18}{cm}.
                
                So far, all reconstructed walls have a thickness of one voxel.
                This results from the fact, that they have been initialised as a contour with the width of one voxel along the border of the ceilings and were then just traced vertically downwards.
                However, also vertical walls as presupposed by this algorithm can encompass elements with a certain extent perpendicular to the wall surface, e.g. applications such as bordures, light switches, window ledges or window recesses that are normally considered as part of the wall and not as individual furniture objects (which would belong to the class 'Indoor Objects').
                To include suchlike elements in our voxel reconstruction of walls, a wall refinement procedure is applied for each pre-existing wall voxel (class 'Wall' or 'Wall Opening').
                New wall voxels arising in the course of this procedure are not recursively processed by it.
                
                First, for each wall voxel, we sweep through the neighbouring voxels along the reversed normal directions of the voxel, i.e. going away from the room to the outside.
                Here, non-empty voxels that do not yet belong to a room are searched within a distance of up to \SI{15}{cm}.
                If there are any non-empty voxels within this search distance, they are added to the wall of the respective room and the normal directions of the initial voxel are copied.
                If there are also empty voxels in-between, these are also added with class label 'Wall Opening'.
                
                Next, we apply a similar procedure going from the initial voxel along its normal directions up to \SI{15}{cm} towards the inside of the room.
                Here, however, we immediately stop if we encounter a voxel labeled as 'Empty Interior'.
                Only a continuous succession of 'Interior Object' voxels can potentially be added to the wall.
                Before doing so, however, we check if this continous succession of voxels goes further than the search distance of \SI{15}{cm}. 
                If this is the case, we assume that they belong to a large furniture object like e.g. a table and do not add them to the wall.
                
            \subsubsection{Wall Opening Refinement}
            \label{sec:method_refinement_wallOpenings}
                
                In a next step, we refine the occurrence of wall opening labels.
                This again is done by traversing the voxel grid along the normal directions towards the room interior and away from it for all pre-existing wall voxels (i.e. all wall voxels that already existed before the last refinement step).
                Thus, for each pre-existing wall voxel, we get stacks of wall voxels along the normal directions.
                If any of these voxels has the label 'Wall', we set all 'Wall Opening' voxels in the stack to 'Wall'.
                
                Furthermore, for all voxels that are still labeled as 'Wall Opening' after applying this refinement step, we again traverse the voxel grid along the normal directions going towards the room interior with a search distance of \SI{70}{cm} to check if there are surfaces of large furniture objects occluding this part of the wall.
                If this is the case, the whole wall voxel stack is set to 'Wall' as well.
        
        \subsection{Evaluation}
        \label{sec:method_evaluation}
        
            As the indoor reconstruction method proposed in this work does not process the geometric primitives comprising the input data-sets themselves but voxel representations derived from them, ground truth data for evaluation purposes should in the end also be given as voxel representation.
            Thus, a possible approach would be to manually label voxel representations of input data with ground truth information.
            However, to be able to automatically evaluate the influence of the voxel size in the scope of future work, we chose not to label our ground truth data in voxel space but to construct ground truth data in the metric space of the input data and pass it through the voxelization step as well.
            
            As our aim is to not only semantically classify the geometric primitives comprising the input data, but also to geometrically reconstruct indoor environments in the voxel representation, generation of ground truth data must exceed simple labelling of the geometric primitives.
            The input data in the form of triangle meshes may contain gaps caused by occlusion or incomplete mapping of the represented indoor environments.
            The reconstruction algorithm, however, is expected to fill these gaps by labelling the voxels there as room surface.
            Thus, suchlike gaps must be manually closed by geometries in the ground truth data.
            
            We hence approached ground truth data generation by manually cutting the triangle meshes comprising the input data in different parts and labelling them.
            First, the meshes are manually split in rooms and then on per-room basis split in semantic classes as 'Floor', 'Ceiling' and 'Wall'.
            The resulting meshes are then completed by manually constructing planes to close gaps in the data, e.g. where walls are occluded by furniture.
            Furthermore, openings in the meshes that should explicitly be detected as openings by the reconstruction algorithm such as window openings or door openings are also manually constructed as planes. 
            
            The resulting labelled ground truth meshes are then voxelized as described in \autoref{sec:method_dataPreparation}, while their labels are passed on to the resulting voxels.
            For evaluation purposes, test data as well as ground truth data are both voxelized by the same grid, i.e. for determining the grid extensions, we use a bounding box including both data-sets.
            Thus, we ensure that voxel coordinates are directly comparable.
            
            Some rules are applied to handle situations, where ground truth meshes with conflicting labels are intersecting the same voxel.
            For instance if a voxel intersects 'Wall' meshes as well as 'Wall Opening' meshes, it should be labelled as 'Wall'.
            Furthermore, the ground truth voxel model should use the same data representation as the one created by the algorithm to evaluate, i.e. voxels can belong to multiple classes and multiple rooms.
            
            As the manual labelling of furniture as 'Interior Object' and the construction of volumetric geometries representing 'Empty Interior' would be too time-consuming, these classes were excluded from the evaluation.
            Voxels belonging to these classes are for evaluation purposes treated as empty voxels not belonging to any room, as they are represented in the ground truth data.
            
            As the labelling of room numbers may differ between test data and ground truth data and, furthermore, the segmentation of rooms may differ between the data-sets, a mapping between rooms of both data-sets has to be determined.
            On the one hand, this serves to evaluate the room segmentation of the proposed algorithm and, on the other hand, this is used for evaluating the voxel classification.
            
            To this end, a weighted mapping between rooms from both data-sets is derived by analysing voxels, that in both data sets are classified as 'Ceiling'.
            If such a voxel is assigned in the ground truth data-set as belonging to room $X$ and in the test data as belonging to room $Y$, the weight counter of room mapping $X \rightarrow Y$ is incremented.
            Room mappings with negligible weight can be discarded.
            
            In the evaluation procedure, we currently only check for corresponding voxels to hold exactly similar states in due consideration of the above-mentioned room mapping and the treatment of room interior voxels as not belonging to a room.
            This means that we do currently not consider cases, where a voxel has the correct state for one room but an incorrect one for another room, as partly correct.
            
\section{RESULTS}
\label{sec:results}

\begin{figure}
    \centering
    \subfigure[Triangle mesh.] {
        \label{fig:attic_mesh}
        \includegraphics[width=8cm]{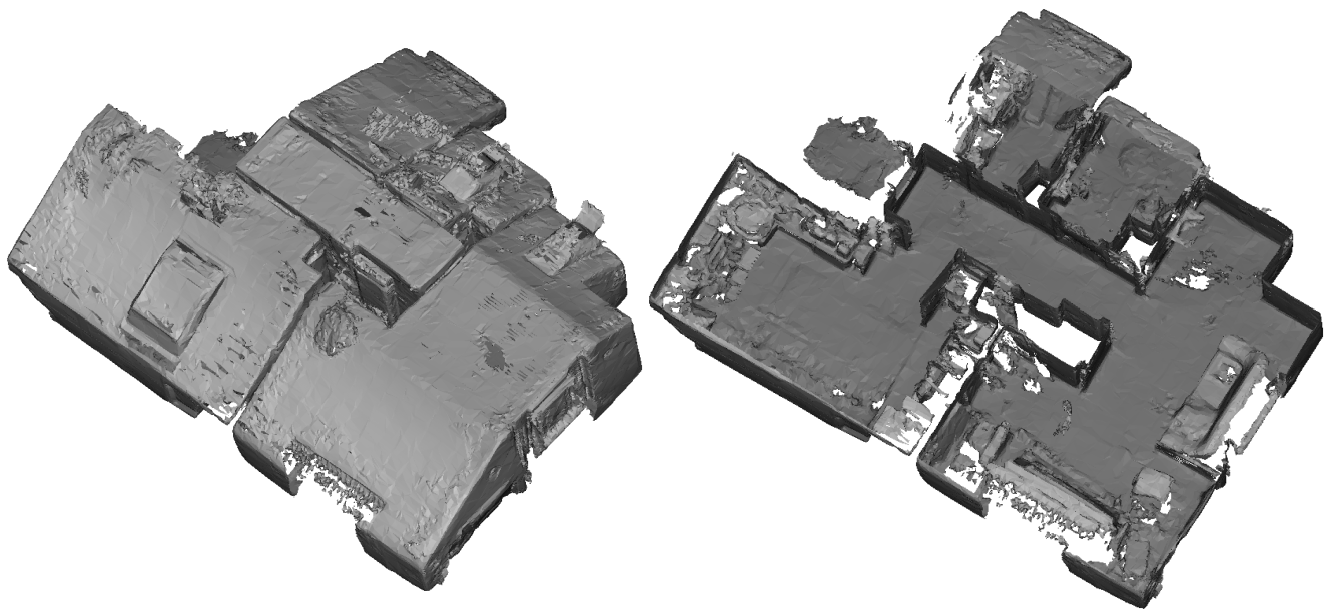}
    }
    \subfigure[Reconstructed voxel model, where voxels are colourised according to Table~\ref{tab:voxelclasses_reconstruction}(b).] {
        \label{fig:attic_reconstruction}
        \includegraphics[width=8cm]{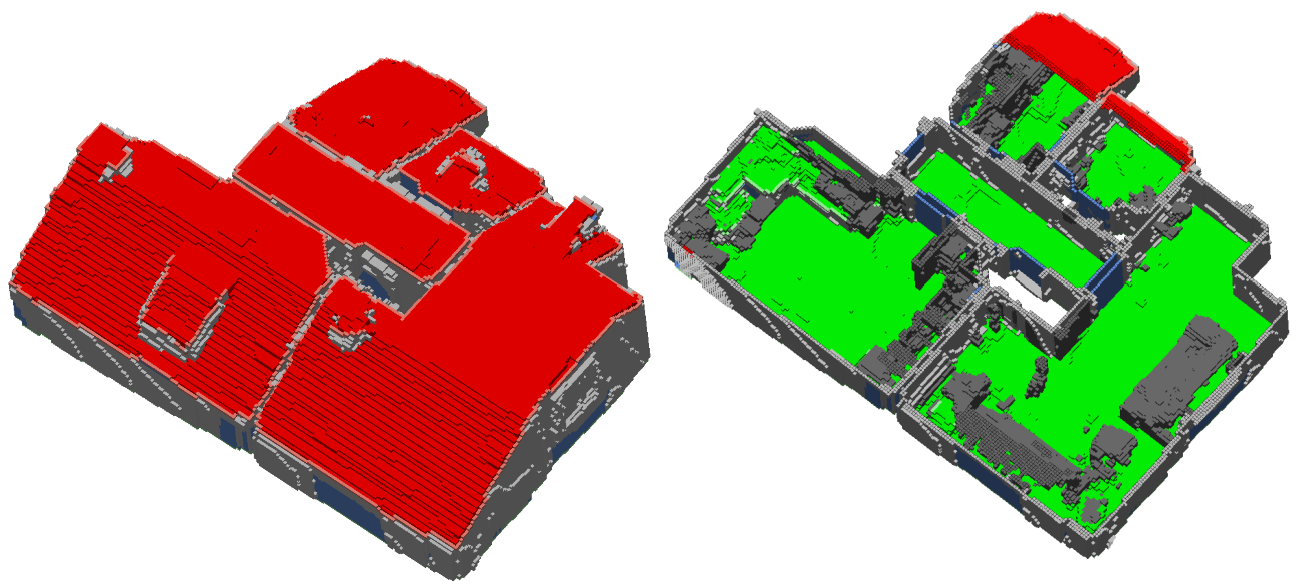}
    }
    \caption { The data set 'Attic'. }
    \label{fig:attic}
\end{figure}

In the evaluation process, we use two data-sets. 
Both triangle meshes were captured with a Microsoft HoloLens (version 1).
The first data-set ('Office') represents an indoor office environment with multiple rooms on two storeys including furniture.
The input mesh as well as the reconstructed voxel model including voxels of the class 'Interior Object' are depicted in \autoref{fig:workflow}.
The second data-set ('Attic') represent an attic with slanted ceilings comprised of five rooms used as storage area.
The triangle mesh and the resulting voxel reconstruction are depicted in \autoref{fig:attic}.

\begin{figure}
    \centering
    \subfigure[Ground truth for the axes-aligned data-set.] {
        \label{fig:voxelized_ground_truth}
        \includegraphics[width=8cm]{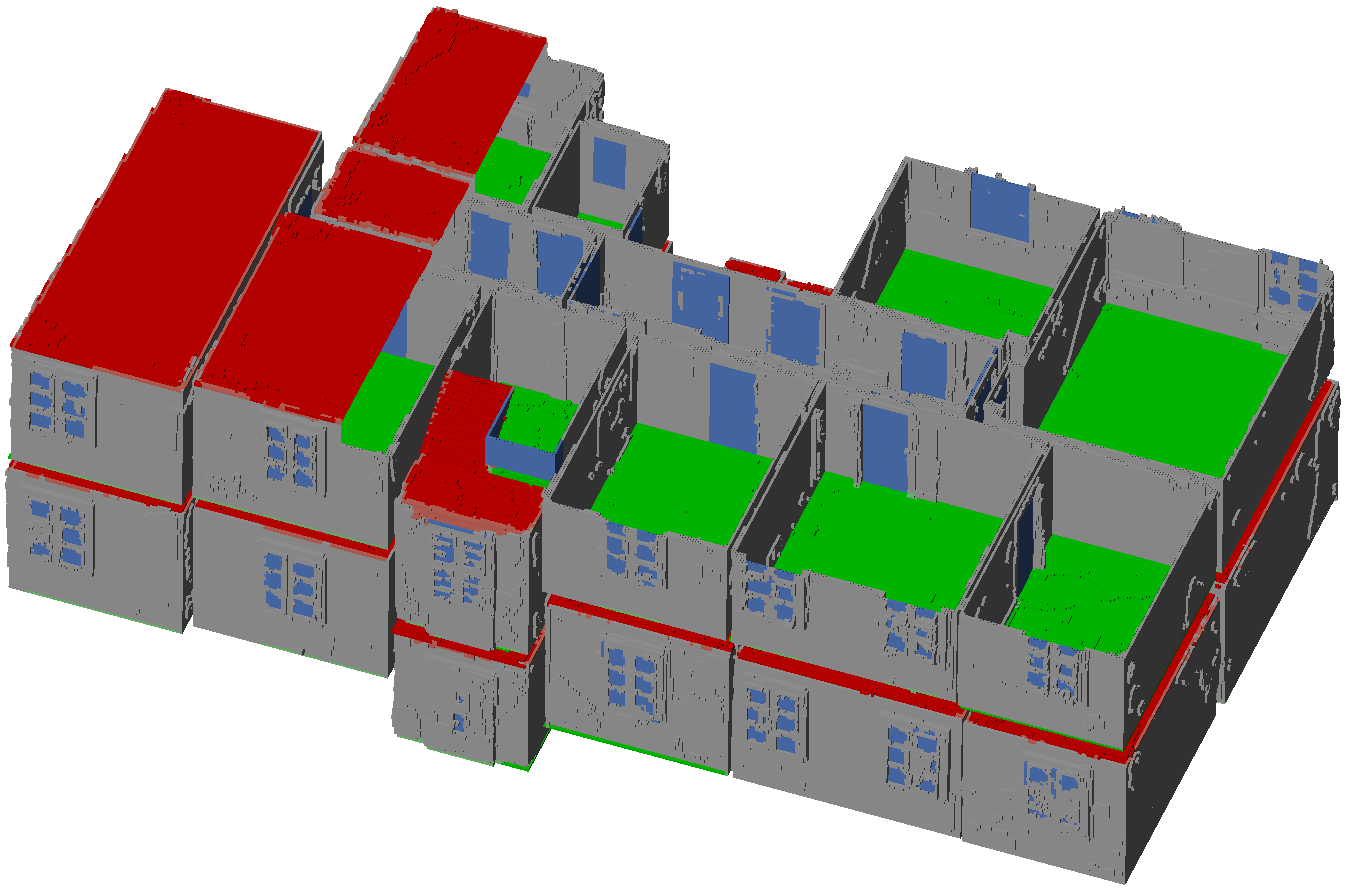}
    }
    \subfigure[Reconstruction result for the axes-aligned data set.] {
        \label{fig:voxelized_reconstruction}
        \includegraphics[width=8cm]{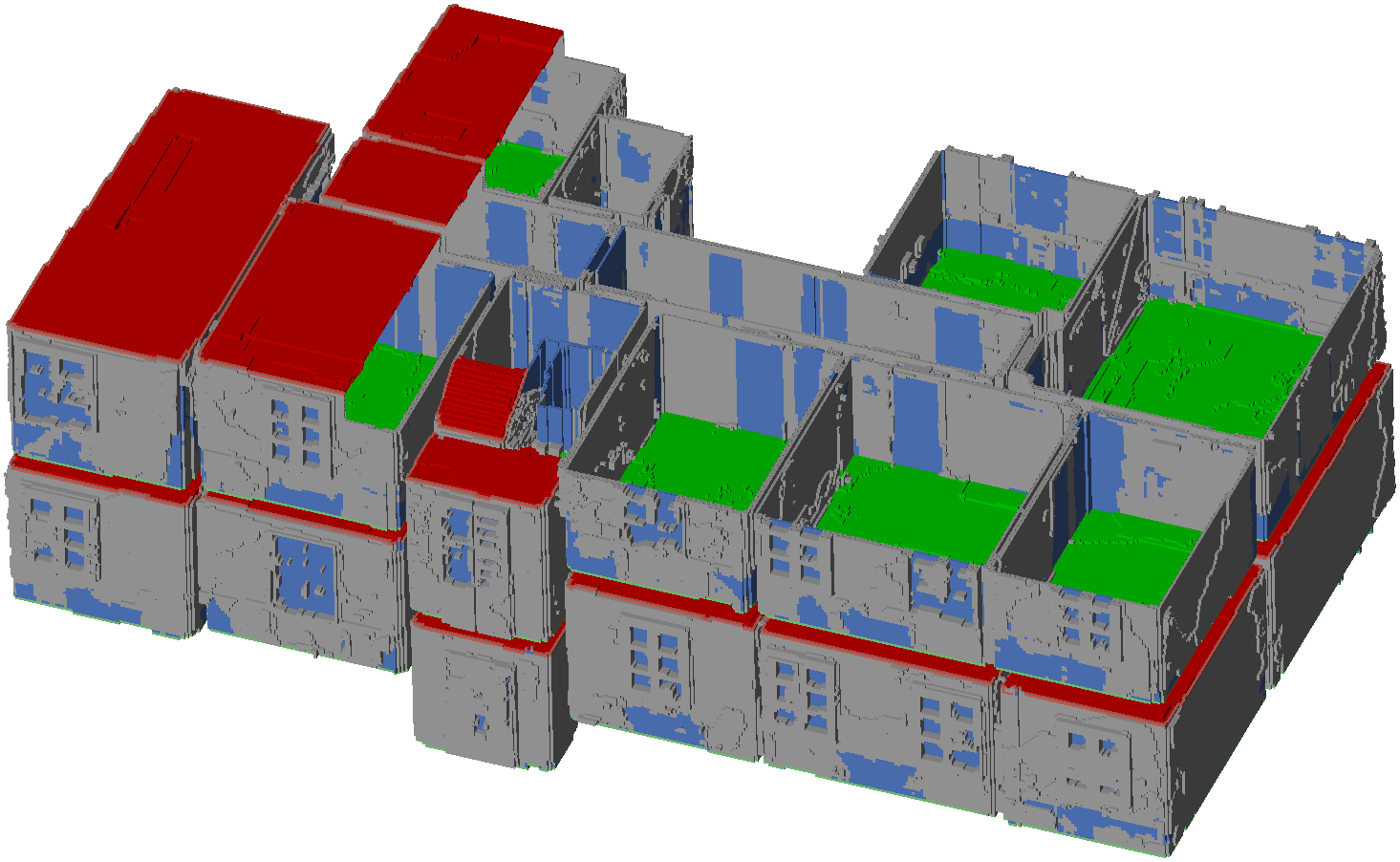}
    }
    \subfigure[Ground truth for the data-set rotated by \SI{30}{\degree} around the up-axis.] {
         \label{fig:voxelized_ground_truth_skewed}
         \includegraphics[width=8cm]{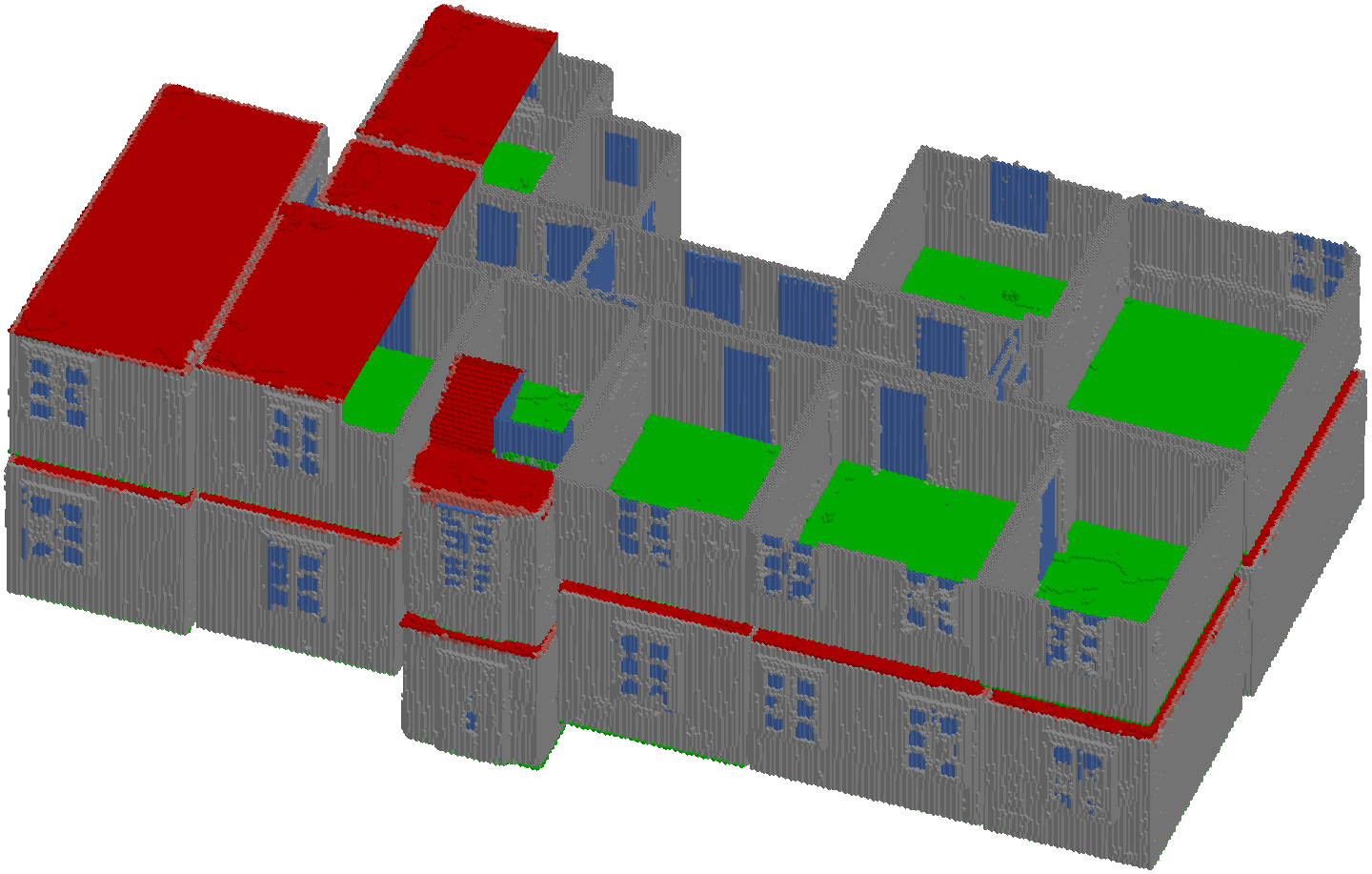}
    }
    \subfigure[Reconstruction result for the data-set rotated by \SI{30}{\degree} around the up-axis.] {
        \label{fig:voxelized_reconstruction_skewed}
        \includegraphics[width=8cm]{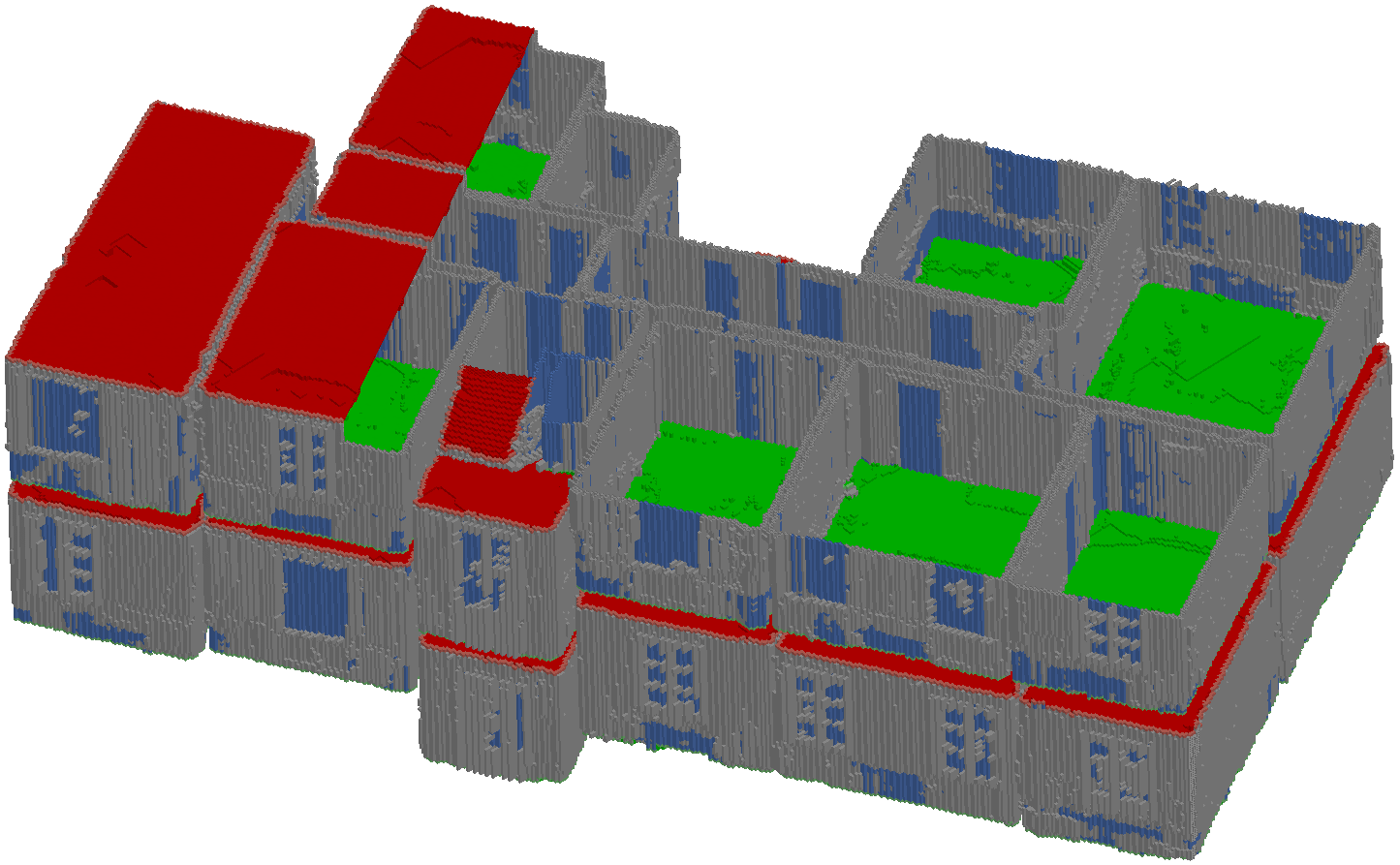}
    }
    \caption { 
        Voxelized ground truth data and reconstruction results for the 'Office' data-set. Voxels are colourised according to Table~\ref{tab:voxelclasses_reconstruction}(b). In each case, a part of the ceiling is removed for the sake of visibility.
    }
    \label{fig:voxelized}
\end{figure}

For the 'Office' data-set, a ground truth voxel model derived from manually labeled and constructed ground truth meshes is depicted in Figure \autoref{fig:voxelized_ground_truth}, while Figure \autoref{fig:voxelized_reconstruction} depicts our reconstruction with voxels of the classes 'Interior Object' and 'Empty Interior' omitted.
To demonstrate that our reconstruction algorithm is not restricted to Manhattan World scenarios, we also rotated the input mesh as well as the ground truth mesh by \SI{30}{\degree} around the up-axis.
The resulting voxel models are depicted in Figures \ref{fig:voxelized_ground_truth_skewed} and \ref{fig:voxelized_reconstruction_skewed}, respectively.

The evaluation results for these data-sets are detailed in \autoref{tab:results}.
In both cases of the 'Office' data-set (axes-aligned and rotated mesh), all rooms except for the stairwell were detected correctly.
The stairwell being defined as one room in the ground truth model was in both cases split into multiple rooms by the reconstruction algorithm.
In the case of the 'Attic' data-set, all five rooms were correctly detected.

For all data-sets, the high overall ratios of correctly classified voxels is heavily biased by an enormous predominance of correctly classified empty voxels.
The low ratios of correctly classified voxels in proportion to non-empty voxels in the reconstructed model, however, is also misleading as will be discussed in the following section.
The ratio of correctly classified voxels in proportion to voxels that are not empty in the reconstruction as well as in the ground truth, on the other hand, is again over 90\,\% for the 'Office' data-set and around 85\,\% for the 'Attic' data-set.
The results for the axes-aligned and the rotated version of the 'Office' data-set are quite comparable, both visually as well as concerning the evaluation results presented in \autoref{tab:results}.

{\newcommand*{\thead}[1]{\multicolumn{1}{c}{\bfseries #1}}
\begin{table}
	\centering
	\caption{
	    Evaluation results (Vx.: Voxels, NE: Non-Empty, GT: Ground Truth, RC: Reconstruction).
	}
	\begin{tabular}{lrrr}
		\toprule
		\thead{Data-Set} & \thead{Office}
		    & \begin{tabular}{@{}c@{}}\textbf{Office}\\\textbf{Rotated}\end{tabular} 
		    & \thead{Attic}  \\
		\midrule
		\textbf{Mesh Vertices} [*10\textsuperscript{6}] & 2.87 & 2.87 & 0.08 \\
		\textbf{Vx.} [*10\textsuperscript{6}]            & 18.89 & 30.26 & 1.73  \\
		\textbf{NE Vx. in RC} [\%]                & 5.24 & 3.23 & 6.8  \\
		\textbf{Rooms GT}                               & 25 & 25 & 5  \\
		\textbf{Rooms RC}                               & 30 & 29 & 5  \\
		\textbf{Wrong Rooms from GT}                    & 1 & 1 & 0  \\
		\textbf{Correct Vx.} [\%]                       & 95.18 & 96.72 & 92.91  \\
		\begin{tabular}{@{}l@{}}\textbf{Correct Vx. in}\\\textbf{RC NE} [\%]\end{tabular} & 51.77 & 53.70 & 44.04  \\
		\begin{tabular}{@{}l@{}}\textbf{Correct Vx. in}\\\textbf{GT and RC NE} [\%]\end{tabular} & 91.66 & 91.82 & 84.53  \\
		
		\bottomrule
	\end{tabular}
	\label{tab:results}
\end{table}
}

\begin{figure}[t]
    \centering
    \subfigure[All cases as listed in \autoref{tab:voxelclasses_errors}.] {
      \includegraphics[width=8cm]{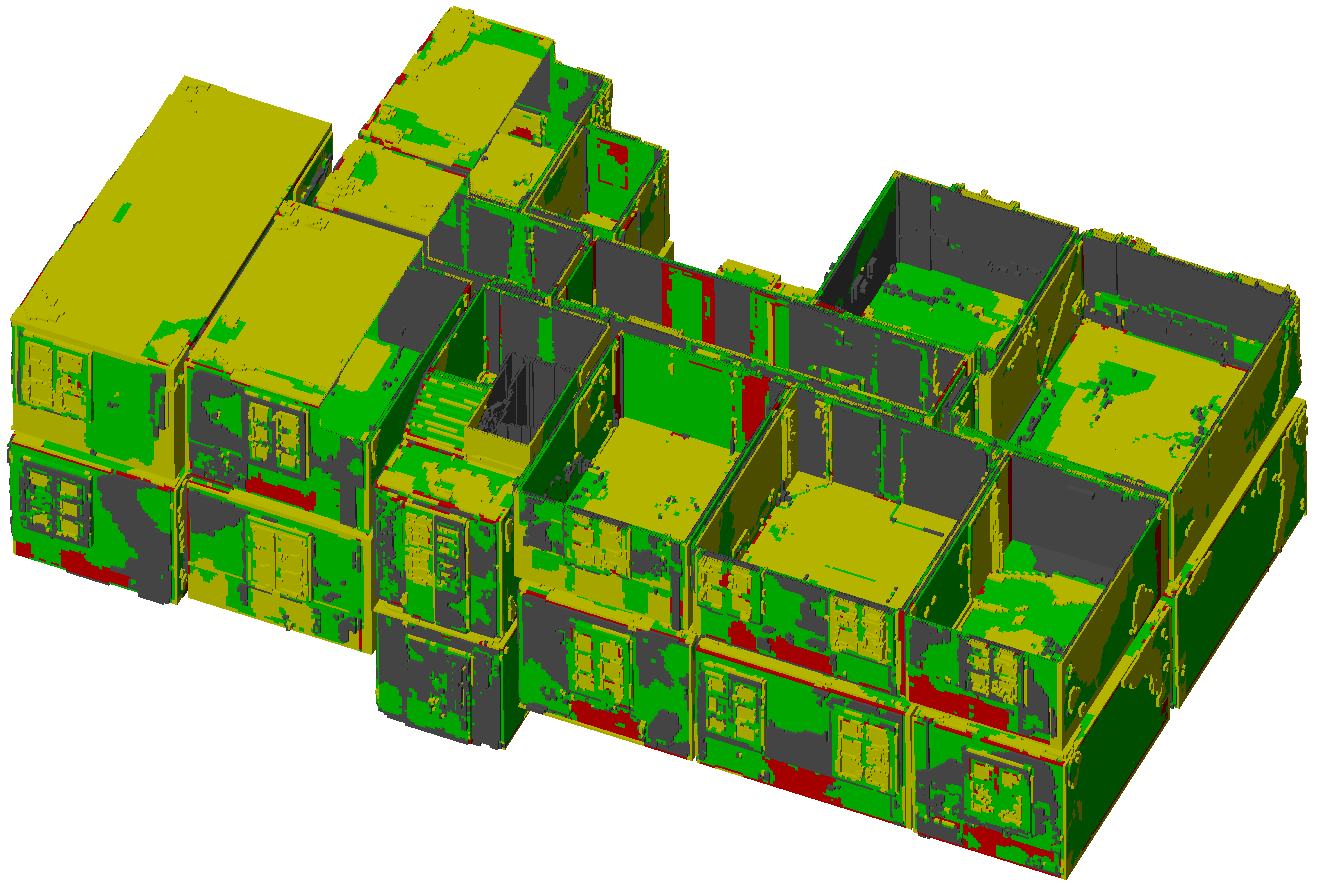}
    }
    \subfigure[Only voxels that are not empty in the reconstruction as well as in the ground truth model.] {
      \includegraphics[width=8cm]{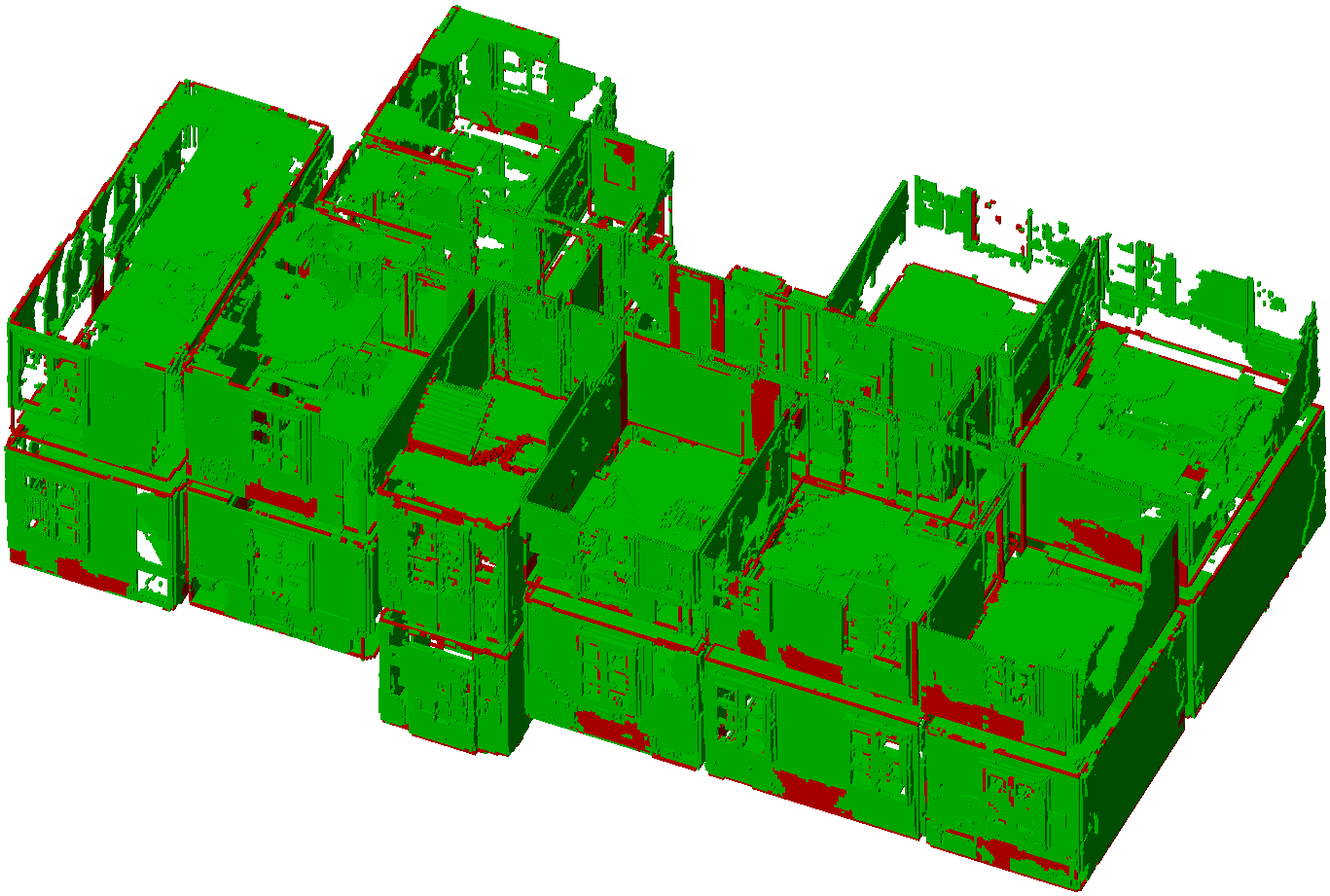}
      \label{fig:error_distribution_b}
    }
    \caption { 
        Spatial distribution of correctly and wrongly classified voxels for the reconstruction of the axes-aligned model from Figure \autoref{fig:voxelized_reconstruction}. The voxels are colourized according to \autoref{tab:voxelclasses_errors}. A part of the ceiling is removed for the sake of visibility.
    }
    \label{fig:error_distribution}
\end{figure}

\autoref{fig:error_distribution} gives an impression of the spatial distribution of wrongly classified voxels for the axes-aligned 'Office' data-set.
Voxels being correctly identified as non-empty but having the wrong class label are mainly constituted by wall voxels missing a second class label for ceiling or floor and false wall openings that should have been classified as 'Wall'.
Furthermore, it is apparent, that the low ratio of correctly classified voxels in relation to non-empty voxels in ground truth or reconstruction is mainly caused by layers of missing or superfluous voxels along the room surfaces.

\begin{table}
    \caption{The meaning of the colours in \autoref{fig:error_distribution} (GT: Ground truth, RC: reconstruction)}.
    \begin{tabular}{cl}
        \toprule
		Color & Label \\ \midrule
		\textcolor{VoxelFloor}{\faCube} & Voxel classified correctly \\
		\textcolor{VoxelCeiling}{\faCube} & Voxel not empty in GT and RC, but wrong class \\
		\textcolor{VoxelInteriorObject}{\faCube} & Voxel empty in GT, but not in RC \\ 
		\textcolor{Yellow}{\faCube} & Voxel empty in RC, but not in GT \\ 
		\bottomrule
	\end{tabular}
	\label{tab:voxelclasses_errors}
\end{table}

\section{DISCUSSION}
\label{sec:discussion}

As already mentioned, the low ratios of correctly classified voxels in proportion to the total count of voxels being not empty in the reconstructed model seems to indicate a quite low quality of the classification results.
However, as \autoref{fig:error_distribution} indicates, missing 

voxels in the reconstruction are in large parts situated as layers along room surfaces as is the case with voxels that should be empty according to the ground truth, but are not in the reconstruction.
The high amount of voxels that should be empty according to the ground truth along wall surfaces can be attributed to the geometric refinement of the walls presented in \autoref{sec:method_refinement_wallGeometry} which results in a thickening of the voxel representation of the reconstructed walls.
Layers of missing voxels, on the other hand, mainly occur on top of ceiling and floor voxel surfaces, which indicates that our approach seems to systematically underestimate the height of these horizontal slabs.

Generally, it needs to be discussed, if an evaluation procedure accounting only for the presence or absence of values on exact voxel positions is adequate in a reconstruction scenario such as this.
However, it would rather be more appropriate to try to quantify the displacement between entities such as walls or slabs between ground truth data and reconstruction.
This, however, presupposes a meaningful segmentation of such entities and the correct allocation of them between reconstruction and ground truth.
Generally, a meaningful comparison between different model representations or mapping results representing the same indoor environment is not a trivial matter.
Valuable discussions on this topic can be found e.g. in \citep{chen_et_al_2018c} or \citep{khoshelham_et_al_2018}.

As indicated by Figure \autoref{fig:error_distribution_b}, there is clearly still potential for improvement in the distinction between actual wall openings and the absence of geometric primitives caused by occlusion.
While checking for large occluding furniture objects directly in front of wall opening voxels works quite well in many cases, occlusions from tables where the occluding object is not in front but above the wall opening are not yet accounted for.
Some cases of erroneous wall openings could also potentially be prevented by checking if the adjacent wall of a neighbouring room shows a corresponding opening.
If this is not the case, it could be reasoned that the respective wall opening can be closed.

Furthermore, over-segmentation of rooms does occur in the case of the stairwell in the 'Office' data-set.
This is caused by edges in the ceiling with horizontal normal directions splitting the ceiling surface into multiple segments.
In cases like this, a refinement of the room segmentation could be applied by checking if room candidates are separated by walls consisting primarily of empty voxels and merging them where appropriate.
In order to realise this, as well as for the refinement of wall openings mentioned before, it would be necessary to determine the adjacency of neighbouring rooms in the voxel model on the basis of the individual voxels but also in terms of reasonably defined wall entities comprised of voxels.

\section{CONCLUSION}
\label{sec:conclusion}

In this work, we presented a novel approach for reconstructing models of indoor environments in the form of voxel representations from unstructured three-dimensional geometries like triangle meshes.
This process encompasses the voxelization of the input data.
Rooms are detected in the resulting voxel grid by segmenting connected voxel components of ceiling candidates and extruding them downwards to find floor candidates.
Semantic class labels like 'Wall', 'Wall Opening', 'Interior Object' and 'Empty Interior' are then assigned to the room voxels in-between ceiling and floor by a rule-based voxel sweep algorithm.
Finally, the geometry of the detected walls is refined in the voxel representation.

We demonstrated by means of quantitative evaluation, that the presented algorithm holds potential for the reconstruction of indoor environments from sparse and noisy data of complicated and clutter-rich three-dimensional indoor environments.
Further work should be directed in merging over-segmented rooms and extracting three-dimensional topological relations between the reconstructed rooms such as adjacency and accessibility through sufficiently large wall openings.
Furthermore, the conversion of the resulting voxel model into more common forms of building data representation such as surface models or volumetric objects would enable the further processing of the reconstructed models with prevalent BIM tools.

A further goal for future research is to make these indoor models derived from data acquired by a mobile augment reality device accessible in the context of indoor augmented reality by enriching them with additional space-related information that can then be visualised directly in the location that it refers to.
An open research question in this context is how a mobile AR device can be automatically localised within large-scale indoor environments only on the basis of an abstracted indoor model without relying on artificial markers or other kinds of infrastructure.

{
    \begin{spacing}{0.7}
        \bibliography{bibliography.bib} 
    \end{spacing}
}

\end{document}